\documentclass[sigconf, nonacm]{acmart}
\pdfoutput=1

\usepackage{amsmath}
\usepackage{graphicx}
\usepackage[textsize=tiny,colorinlistoftodos]{todonotes}
\newcommand{\squishlist}{
   \begin{list}{$\bullet$}
    { \setlength{\itemsep}{0pt}      \setlength{\parsep}{3pt}
      \setlength{\topsep}{3pt}       \setlength{\partopsep}{0pt}
      \setlength{\leftmargin}{1.5em} \setlength{\labelwidth}{1em}
      \setlength{\labelsep}{0.5em} } }

\newcommand{\squishend}{
    \end{list}  }

\begin{document}

\title{INODE: Building an End-to-End Data Exploration System \\ in Practice [Extended Vision]}


\author{Sihem Amer-Yahia$^2$, Georgia Koutrika$^1$, Frederic Bastian$^7$,  Theofilos Belmpas$^1$,  Martin Braschler$^9$, Ursin Brunner$^9$, Diego Calvanese$^8$, Maximilian Fabricius$^5$, Orest Gkini$^1$, Catherine Kosten$^9$, Davide Lanti$^8$, Antonis Litke$^6$, Hendrik L\"{u}cke-Tieke$^3$, Francesco Alessandro Massucci$^6$,
Tarcisio Mendes de Farias$^7$, Alessandro Mosca$^8$, Francesco Multari$^6$, Nikolaos Papadakis$^4$,  Dimitris Papadopoulos$^4$, Yogendra Patil$^2$, Aurélien Personnaz$^2$,  Guillem Rull$^6$,  Ana Sima$^7$, Ellery Smith$^9$, Dimitrios Skoutas$^1$, Srividya Subramanian $^5$, Guohui Xiao$^8$, Kurt Stockinger$^9$ \\}
\affiliation{
  (1) Athena Research Center, Greece (2) CNRS, University Grenoble Alpes, France (3) Fraunhofer IGD, Germany \\ 
  (4) Infili, Greece (5) Max Planck Institute, Germany (6) SIRIS Academic, Spain \\ 
  (7) SIB Swiss Institute of Bioinformatics, Switzerland (8) Free University of Bozen-Bolzano, Italy \\
  (9) ZHAW Zurich University of Applied Sciences, Switzerland
}

\renewcommand{\shortauthors}{INODE Project Partners}

\begin{abstract}
A full-fledged data exploration system must combine different access modalities with a powerful concept of \textit{guiding} the user in the exploration process, by being \textit{reactive} and \textit{anticipative} both for data discovery and for data linking. Such systems are a real opportunity for our community to cater to users with different domain and data science expertise. 

We introduce INODE - an end-to-end data exploration system - that leverages, on the one hand, Machine Learning and, on the other hand, semantics for the purpose of Data Management (DM). Our vision is to develop a classic unified, comprehensive platform that provides extensive access to open datasets, and we demonstrate it in three significant use cases in the fields of Cancer Biomarker Research, Research and Innovation Policy Making, and Astrophysics. INODE offers sustainable services in (a) data modeling and linking, (b) integrated query processing using natural language, (c) guidance, and (d) data exploration through visualization, thus facilitating the user in discovering new insights. We demonstrate that 
our system is uniquely accessible to a wide range of users from larger scientific communities to the public. Finally, we briefly illustrate how this work paves the way for new research opportunities in DM. 
\end{abstract}

\maketitle

 \section{Introduction}
The Data Management (DM) community has been actively catering to Machine Learning (ML) research by developing systems and algorithms that enable data preparation and flexible model learning. This has resulted in several major contributions in developing ML pipelines, and formalizing algebras and languages to facilitate and debug model learning, as well as designing and implementing algorithms and systems to speed up ML routines~\cite{DBLP:journals/pvldb/BoehmDEEMPRRSST16, DBLP:conf/icde/SparksVKFR17}. 
Existing work that leverages ML for DM~\cite{DBLP:conf/sigmod/Stoica20} is nascent and covers the use of ML for query optimization~\cite{DBLP:conf/sigmod/KristoVCMK20} or for database indexing~\cite{DBLP:conf/sigmod/KraskaBCDP18}. 
This paper makes the case for democratizing \emph{Intelligent Data Exploration} by leveraging ML for DM.

Traditionally, database systems assume the user has a specific query in mind, and can express it in the language the system understands (e.g., SQL).
However, today, users with different technical backgrounds, roles, and tasks are accessing and leveraging voluminous and complex data sources. In many scenarios, they are only partially familiar with the data and its structure, and their needs are not well-formed. In such settings, \emph{expanding traditional query answering to data exploration is a natural consequence and requirement and with it comes the need to redesign systems accordingly}. This need translates to several challenges at different levels.


(\emph{Interaction}). Regarding interaction with the system, the biggest challenge is to enable the user to express her needs through a variety of \textbf{access modalities}, ranging from SQL and SPARQL to natural language (NL) and visual query interfaces, that can be used and intermingled depending on the user needs and expertise as well as the data exploration scenario. 
 The second challenge is that of \textbf{user guidance}, i.e., users should be allowed to provide {\em feedback} to the system, and the system should leverage that feedback to improve subsequent exploration steps.

\begin{figure*}[t]
\includegraphics[width=6.7in]{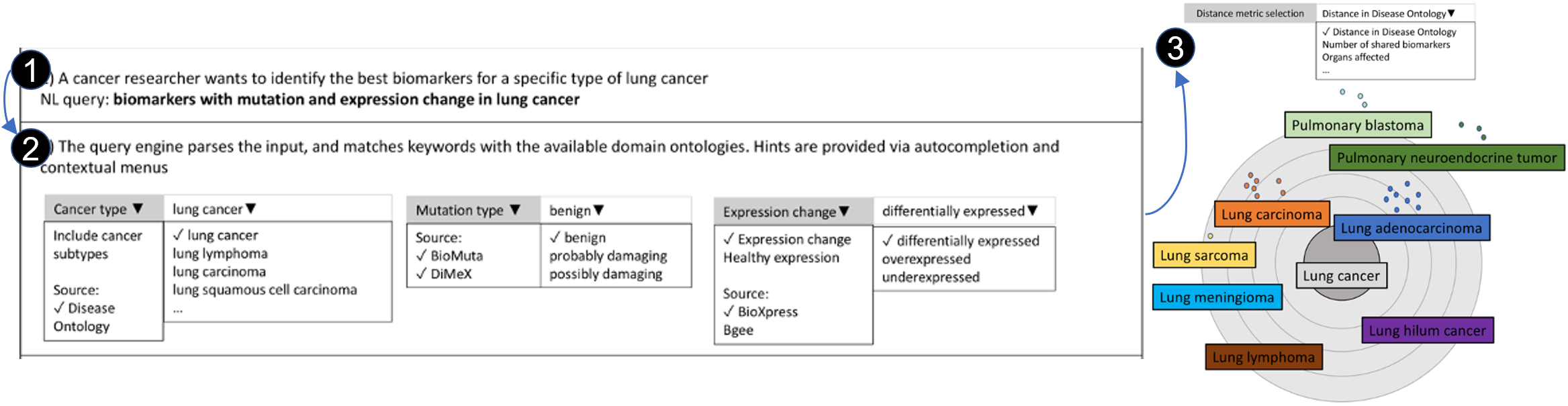}
\caption{Natural language query interface with user assistance. Step 1: User enters a natural language query. Step 2: System parses query and matches keywords against the available ontology to enable term disambiguation; the user iterates the process. Step 3: System visualizes various cancer types that are similar to lung cancer. The distance metric between the diseases can be chosen by the user, e.g. by semantic distance.}
\label{fig:CancerResearch1}
\end{figure*}

(\emph{Linking}). 
Once a user need has been formulated and sent to the system, it is executed over a (fixed) data set. Users may be aware which additional data sets could be of interest. However, they do not always know how to correctly link, integrate, and query more than one data source to generate rich information. This introduces the challenges of \textbf{data linking}, so that new data sources can be added to the system, as well as \textbf{knowledge generation}, so that queries over unstructured data can be supported. Both of these aim at enabling the continuous expansion of the ``pool'' of available data sources, thus making more data available to users.

\emph{(Guidance)}. 
Traditionally, the system will return to the user a set of tuples that concludes the search. There is a lot of work on how to improve performance for query workloads (predict future queries, build indices adaptively, etc.), but still the system has a rather passive role: anticipating or at best trying to predict the next query and then optimize its performance accordingly. Hence, the challenge of \textbf{system proactiveness} arises. The output is not only the set of results but also recommendations for subsequent queries or exploration choices. In our vision, the system guides the user to find interesting, relevant or unexpected data and actively participates in shaping the query workload.

In a nutshell, a full-fledged data exploration system must \emph{combine different access modalities with a powerful concept of guiding the user} in the exploration process. It must {\it be reactive and anticipative} co-shaping with the user the data exploration process. Finally, while data integration has been around for a while, {\em the ability to tie together data discovery and linking is a central question in an intelligent data exploration system.}

\emph{(Evaluation)}. An essential part of our proposal is the development of an evaluation framework to enable the end-to-end assessment of an intelligent data exploration system. This requires to formalize system and human metrics that are necessary for data linking and integration, multi-modal data access, guidance, and visualization.



\textbf{Related Work}. Several systems address components of our vision. A number of them enable NL-to-SQL~\cite{blunschi2012soda},  SQL-to-NL~\cite{logos} or both \cite{DBLP:conf/cidr/JohnPP17} (see a summary in \cite{affolter2019comparative}).  Recommendation strategies can be leveraged to guide users \cite{DBLP:conf/ssdbm/LiuZPM19}. Work on interactive data exploration aims at helping the user discover interesting data patterns based on an integration of classification algorithms and data management optimization techniques \cite{DBLP:journals/tkde/DimitriadouPD16}. Each of the above-mentioned systems tackles specific data management challenges as so-called \textit{insular solutions}. However, these insular solutions have not been integrated to tackle the end-to-end aspect of intelligent data exploration targeted at a wide range of different end users.

Combining all the challenges above requires an elaborated system whose multi-aspect behavior and functionality is the result of a synergy between disjoint technologies, and integrates them into a new ensemble.
This gives rise to multiple approaches that vary in computational complexity, and raises new challenges that can benefit from recent advances in ML.



\textbf{In summary},  this paper makes the following contributions.
We advocate for using ML to solve DM problems that arise when building intelligent data exploration systems. We illustrate the need for intelligent data exploration with relevant use cases (Section~\ref{sec:usecases}). We  describe the solution we have today, INODE\footnote{http://www.inode-project.eu/}, that we are currently building as part of a European project (Section~\ref{sec:today}). To fully complete our vision, we provide open research challenges to be addressed  at the intersection of DM and ML (Section~\ref{sec:tomorrow}).

\section{Use Cases}
\label{sec:usecases}

In this section, we describe two use cases from cancer research and astrophysics and show how INODE can tackle them.


\vspace{4pt}
\underline{Use Case 1: Cancer Research} (Natural Language and Visual Data Exploration).
Fred is a biologist who studies cancer. His goal is to find which specific biomarkers are indicators for a certain type of lung cancer. He needs natural language exploration. 

INODE offers support for NL queries, query recommendations, and interactive visualizations triggered by NL queries (see Figure \ref{fig:CancerResearch1}).
For instance, Fred starts with \textbf{a request in NL} for the topics related to lung cancer but is not sure how to continue after inspecting the results. 
INODE steps up and \textbf{recommends different options}: to expand the search using experimental drugs for treating lung cancer, or to focus on a subset of lung cancer types associated with a certain gene expression. Fred chooses to \textbf{expand his search} to one of the recommended topics, and receives a new list of lung cancers, drugs and genes. Additionally, INODE \textbf{explains in NL} how results are related. That helps him in selecting experimental drugs for certain gene expressions.  After a few such queries, the system \textbf{visually analyzes the results} for Fred to study. Fred learns about the similarity between different types of cancer based on distance metrics that he can choose.


\vspace{4pt}
\underline{Use Case 2: Astrophysics} (Exploration with SQL-Pipelines).
\label{sec:astro}
In the era of big data, astronomers need to analyze dozens of databases at a time. With the ever increasing number of publically available astronomical databases from various astronomical surveys across the globe, it is becoming increasingly challenging for scientists to penetrate deep into the data structure and their metadata in order to generate new scientific knowledge. Sri, an astrophysicist, explores astronomical objects in SDSS, a large sky survey database\footnote{https://www.sdss.org/}. Sri would like to examine Green Pea galaxies, first discovered in a citizen science project called 'Galaxy zoo', that recently gained attention in astronomy as one of the potential sources that drove cosmic reionization. 

\begin{figure}[t]
\includegraphics[width=\columnwidth]{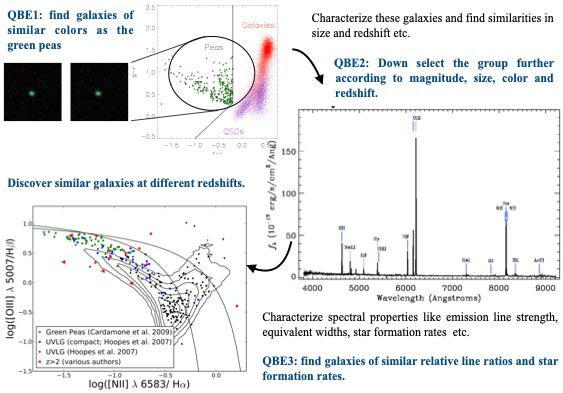}
\caption{Exploring astrophysics data.}
\label{fig:astro}
\vspace{-5pt}
\end{figure}

Figure~\ref{fig:astro} shows a sequence of three consecutive processes of analyzing astrophysics data. Sri relies on selected examples at each step and requests to see comparable ones. In the first query, she \textbf{asks to find galaxies with similar colors} as Green Pea galaxies, she \textbf{requests objects with similar spectral properties, like emission line measurements, star formation rates etc.,} as those returned in the first step. The last query finds \textbf{similar galaxies in terms of their relative ratios and strength of emission lines}. As a result, Sri discovers that green pea emission line ratios are similar to high redshift galaxies.

INODE guides any user in making such new discoveries in an intuitive simpler way, without having to write complicated SQL queries or manual analysis of thousands of galaxies. For instance, INODE helps a user {\bf choose among similarity dimensions} rather than rely on her ability to provide them. Additionally, INODE shows to the user {\bf alternative queries} to pay attention to, thus increasing the chances of making new discoveries.



\section{Current INODE Architecture}  
\label{sec:today}

\begin{figure}[t]
\includegraphics[width=3in]{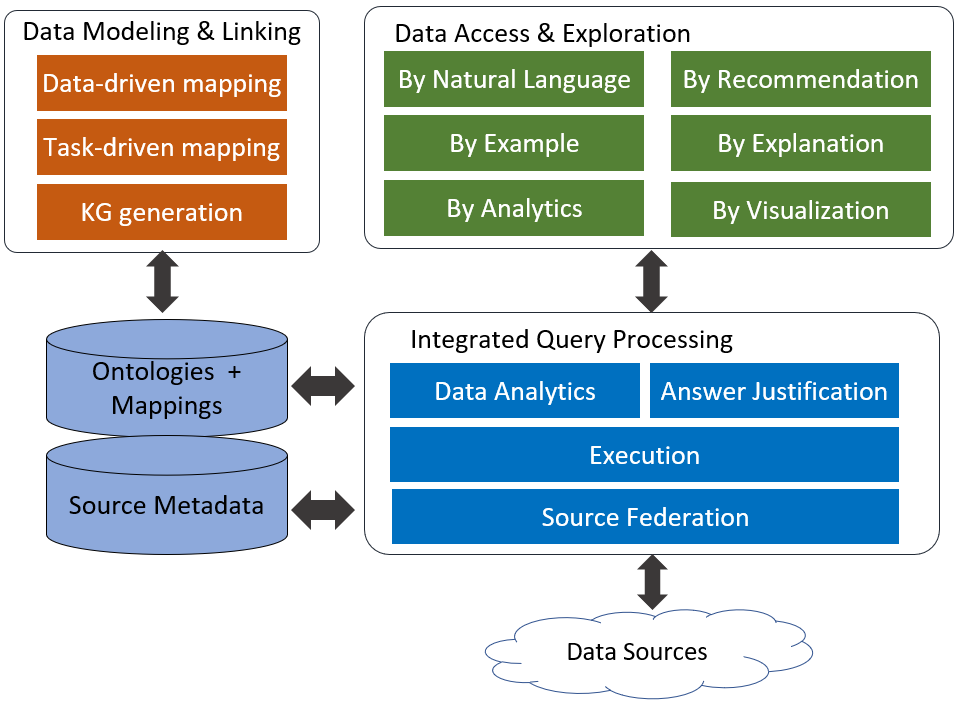}
\caption{Major components of the INODE architecture.}
\label{fig:INODE_architecture}
\vspace{-12pt}
\end{figure}

The main novelty of INODE is bringing together different data management solutions to enable \textit{intelligent data exploration} (see
Figure~\ref{fig:INODE_architecture}). Although some of these solutions and research challenges have been tackled previously, they have not been combined into such an end-to-end intelligent data exploration system, which in turn opens up new research challenges.

INODE's major components are as follows: (1) \emph{Data Modeling and Linking} enables integration of both structured and unstructured data. (2) \emph{Integrated
 Query Processing} enables efficient query processing across federated databases leveraging ontologies. (3) \emph{Data
 Access and Exploration} enables guided data exploration in various modalities such as by natural language, by certain operators or visually.
%

\subsection{Data Modeling and Linking}

This component links loosely coupled collections of data sources such as relational databases, graph databases or text documents based on the
well-established \emph{ontology-based data access} (OBDA)
paradigm~\cite{XCKL*18}. OBDA uses a global ontology (knowledge graph) to \emph{model} the
domain of interest and provides a conceptual representation of the information in the data sources. The sources are linked to elements in
the global ontology through declarative \emph{mappings}. 
It is well-known that designing OBDA mappings manually is a time-consuming and error-prone task. The Data Modeling and Linking component of INODE aims at automatizing this task by providing two mechanisms: {\it data-driven} and {\it task-driven mapping} generation.

\vspace{3pt}
\emph{Data-driven Mapping Generation}.
This mechanism deals with linking novel data sources to the system.  The idea
is to rely on \emph{mapping patterns} that describe well-assessed and sound
schema-transformation rules usually applied in the design process of relational
databases. By analyzing (driven by the patterns) the data
sources, it is possible to automatically derive a so-called \emph{putative
 ontology}~\cite{SeMi15} describing both the explicit entities and
relationships constituting the schema and the implicit ones inferrable from the data. From the mapping patterns, one can also automatically derive
mappings that link the data sources to the putative ontology.

\vspace{3pt}
\emph{Task-driven Mapping Generation}.
This mechanism is applied whenever a task or a query is formulated that uses specific
\emph{target} ontology elements that are not yet aligned with the putative
ontology. In such scenario, the semantics of the query are used to automatically generate mappings
to align the target ontology with the putative ontology.

\vspace{3pt}
\emph{Knowledge Graph (KG) Generation}.
This service transforms unstructured information hidden in large quantities of text (e.g. repositories of scientific papers) to an exploitable structured representation through an NLP pipeline.  INODE follows an Open Information Extraction (OIE) approach to convert each sentence of the corpus into a set of relational triples, where each triple consists of a subject, an object, and a predicate (relationship) linking them. We leverage a number of preprocessing techniques, including co-reference resolution and extractive summarization to improve the quality of the extracted relational triples. We combine different OIE methods (rule-based, analytics-based and learning-based) to achieve both high precision and high recall \cite{app10165630}. The relational triples are further linked with domain-specific ontology concepts before being integrated into the knowledge graph.

\subsection{Integrated Query Processing}

This component is responsible for the execution of queries using 
Ontop~\cite{ISWC-2020-ontop}, a the state-of-the-art OBDA system. Ontop allows the users to formulate queries in terms of \textit{concepts}
and \textit{properties} of their domain of expertise (represented in \textit{knowledge graphs}), rather than in terms of table and attribute names used in the actual data sources. Hence, users do not have to be
aware of the specific storage details of the underlying data sources in order to satisfy
their information needs. 

\vspace{3pt}
\emph{Query Execution}.
This service provides on-the-fly \emph{reformulation} of SPARQL queries over
the domain ontology to SQL queries over the data sources. An approach
based on reformulation has the advantage that the data available in the data
sources does not need to be duplicated in the query processing system, but can
be kept in the data sources as-is. This means that the Query Execution service
is guaranteed even in the common scenario where the user does not own the data
nor does have the right to copy them. To produce reformulations that can efficiently
be executed over the data, in INODE we use optimization techniques such as
\emph{self-join elimination for denormalized data}~\cite{ISWC-2020-ontop} and
\emph{optimizations of left-joins} arising from \texttt{OPTIONAL} and
\texttt{MINUS} operators~\cite{XKCCB18}. 


\vspace{3pt}
\emph{Source Federation}.
The Source Federation service deals with distributing the processing of queries
over the available data sources. 
 INODE provides different kinds of federation
ranging from SQL federation to seamless SPARQL federation. With respect to SPARQL, we
can distinguish between two forms of federation: \emph{seamless federation} and
SPARQL~1.1 \texttt{SERVICE} federation.

In seamless federation, users send queries against a unified view of the remote
endpoints without the need to be aware of the actual vocabularies used in the
federated endpoints. The challenge is to automatically detect to which sources which
components of the query need to be dispatched, to collect the retrieved
results, and to combine them into a coherent answer. We address this challenge
by relying on the knowledge about the sources encoded in the OBDA mappings.
Given that efficiency is a crucial requirement in this setting, our approach requires
a dedicated cost-model able to minimize the number of distributed joins over the federation layer,
in order to favor more efficient joins at the level of the sources.

Instead, \texttt{SERVICE} federation might be adopted in those cases where
users want to directly refer to ``external'' SPARQL endpoints, not yet
integrated with the ontology.  In this setting, the user directly references
the desired endpoints at query time using the SPARQL~1.1 \texttt{SERVICE}
functionality. Observe that the user is required to be aware of the vocabulary
used in the external endpoint.  Hence, the Source Federation service can simply
delegate the execution of the SPARQL query component referenced in the
\texttt{SERVICE} call.

\vspace{3pt}
\emph{Data Analytics}.
The data analytics service exploits novel and efficient query reformulation and
optimization techniques~\cite{ISWC-2020-ontop} to compute complex analytical
functions.  Such techniques are based on algebraic transformations of the
SPARQL algebra tree, rather than on Datalog transformations as traditionally
done in the OBDA literature. This shift of paradigm allows for an efficient
implementation of analytical functions such as SPARQL aggregates. It is worth
noting that INODE, through Ontop, provides the first open-source
reformulation-based system able to support SPARQL aggregates.

The Answer Justification service generates in an automatic way compact and easy
to understand explanations for query results.  In an OBDA setting, the
explanations for a result must take into account, in addition to the query, the
three components of the input, namely the ontology, the mappings, and the data
sources~\cite{IJCAI-2019}.  The ontology is taken care of by identifying the
ontology axioms used for the rewriting of the input query.  As for the
mappings, those considered in the justification are the ones that were used for
the unfolding of the rewritten query to produce the SQL reformulation. Finally,
the data is taken care of by identifying the actual tables and tuples that
contributed to build the considered result, using an approach based on
provenance semirings~\cite{SJMR18}.



\subsection{Data Access and Exploration}

\vspace{3pt}
\emph{Exploration by Natural Language}.
For translating a natural language question into SQL or SPARQL, INODE uses {\it pattern-based}, {\it graph-based} and {\it neural network-based approaches}. For translating from NL to SQL, INODE extends the pattern-based system SODA \cite{blunschi2012soda} with NLP techniques such as lemmatization, stemming and POS tagging to allow both key word search queries as well as full natural language questions. In addition, we use Bio-SODA \cite{sima2019enabling}, a graph-based system to enable NL questions over RDF graph databases. In order to enable federated queries across both relational databases and RDF graph databases, INODE uses an ontology-based data access technology leveraging Ontop \cite{calvanese2017ontop}. The advantage of using pattern-based or graph-based approaches over neural network-based approaches is that they do not require training data, which is often very costly to gain.

Since neural network-based approaches typically require large amounts of training data, we also experimented with various training data generation approaches. INODE uses an {\it inverse data annotation approach} called OTTA \cite{deriu2020methodology}. Rather than writing NL questions and then the corresponding SQL or SPARQL statements, OTTA reverses the process. In particular, OTTA randomly generates so-called operator trees which are similar to logical query plans that can easily be understood by non-tech savvy users. Afterwards, given these operator trees, crowd workers write the corresponding NL questions. In INODE, we use both crowd workers and domain experts for generating training data.

Finally, INODE integrates the {\it neural network-based approach} ValueNet, which leverages transformer architectures to translate NL to SQL \cite{brunner2021valuenet}. The ultimate goal of INODE is to combine all these techniques into an {\it intelligent hybrid-approach} that improves on the errors of each of the individual systems.

\vspace{3pt}
\emph{Exploration by Explanation}.
One of the biggest hurdles in today's exploration systems is that the system provides no explanations of the results or system choices. Nor does the system trigger input from the user, for example, by asking the user to provide more information.
In INODE, we enable a conversational setting, where the system can (a) ask clarifications and (b) explain results in natural language. This interaction assumes that the system is capable of analyzing and understanding user requests and generating its answers or questions in natural language. 

\begin{figure*}[htpb]
 \includegraphics[scale = 0.55]{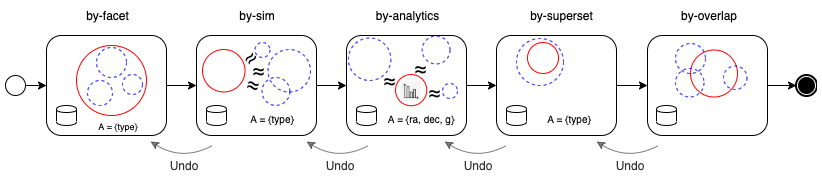}
  \caption{Example data exploration pipeline for analyzing galaxies.}
  \label{fig:sdss}
  \vspace{-10pt}
\end{figure*}

One approach used in INODE builds on {\it Template-based Synthesis} \cite{logos}. This approach considers the database schema as a graph and a query as a subgraph. We use templates that tell us how to compose sentences as we traverse the graph and we use different traversal strategies that generate query descriptions as phrases in natural language.
Furthermore, to generate NL descriptions that use the vocabulary of a particular database, INODE enriches its vocabulary by \emph{leveraging ontologies} built by the Data Modeling and Linking components. 
To further improve INODE's explanation capabilities, we are working on an approach to automatically learn templates, which is especially critical for databases with no descriptive metadata, such as SDSS. Essentially, we are using \emph{neural-based methods} to translate from SQL or SPARQL to natural language.

For example, user questions in natural language inevitably hide ambiguity. Hence, the system can come up with several possible query interpretations that may lead to unexpected results or need the user’s help for query disambiguation. 


\vspace{3pt}
\emph{Exploration by Example and by Analytics}.
By-example is a powerful operator that encapsulates multiple semantics. It takes a set of examples, such as galaxies or patients, and explores its different facets, filters them, finds similar/dissimilar sets, finds overlapping sets, joins them with other sets,  finds a superset, etc. Additionally, by-example operators can be combined with by-analytics to find sets that are similar/dissimilar wrt some value distributions. Figures~\ref{fig:sdss}  shows an example data exploration pipeline (DEP) for exploring galaxies with different instances of by-example.


By-example and by-analytics operators can be represented in the {\it Region Connection Calculus} (RCC)~\cite{LI2003121} and are, in their general form, computationally challenging. For instance, by-subset is akin to solving a set cover problem, which has been extensively studied~\cite{DBLP:conf/cikm/CormodeKW10}. Similarly, by-join requires to have appropriate indices. In INODE, we adopt two approaches. One is based on a {\textit relational backend} in which individual operators are translated into SQL. The other one is an {\textit in-memory Python} implementation that relies on pre-computing and indexing sets. For some operators such as by-facet, the SQL version is straightforward since it resolves into generating a \texttt{GROUP BY}-query. For others, such as by-overlap, the Python version is simpler as it relies on using an index that records pairwise overlaps between pre-computed sets.


\vspace{3pt}
\emph{Exploration by Recommendation}.
In a mixed-initiative setting, the system actively guides the user in what possible actions to perform or data to look at next. 
In INODE, we are interested in recommendations in both \emph{cold-start} (where the user has not given any input) and {\it warm-start} settings (where the user has asked one or more queries but may not know what to do next). In the former case, the goal is to show a set of example or starter queries that the users could use to get some initial answers from the dataset (e.g. \cite{DBLP:conf/sigmod/HoweCKB11}). In the latter case, the system can leverage the user's interactions (queries) to show possible next queries (e.g., \cite{DBLP:journals/corr/abs-1802-02872}).

A big differentiator is the availability of query logs. In case no query logs are available, the system should still provide recommendations.
In INODE we are addressing the recommendation problem from different angles, i.e., generating recommendations: (a)  based on \emph{data analysis} \cite{PyExplore} (b) by \emph{NL-based processing and query augmentation techniques} leveraging knowledge bases (c) by \emph{user log analysis}. 

\vspace{3pt}
\emph{Exploration by Visualization}.
In information retrieval, search queries result in a list of candidates ranked by their matching score \cite{manningIntroductionInformationRetrieval2008}. This also holds true for INODE, as most exploration operators generate multiple potential answers. However, results are not individual items such as documents, but data sets (i.e. sets of items) and have to be communicated to the user differently to support their goals. 
Not only do users have to decide, which data set contains the answer they are looking for, but also to compare the results, to assess redundancies, discrepancies and other surprising or interesting differences to draw hints on how to continue the exploration.

The goals of the \emph{by-visualization} data access and exploration interface are two-fold: (1) Enable "explorers" to understand, compare and decide based on the provided results and (2) enable them to interact with the results by enabling indirect query manipulation, identifying and highlighting parts that are of interest for further analysis and guiding them towards interesting regions  \cite{Stahnke2016, steigerVisualAnalysisTimeSeries2014, ruotsaloSupportingExploratorySearch2013, may2012using}. Depending on the users information need, the optimal answer may take different forms. For example, some questions can be satisfied by a single data cell, while others require aggregated values or even multiple tables. Hence, making the results explorable with respect to the users' needs is very challenging.

Our processes for user requirements elicitation confirms our goals stated above and is based on the {\it User Centered Design} standard \cite{ISO9241-UCD}. 
In addition to that, users emphasized the importance to compare differences as well as similarities of queries and results.
As a base line, we enabled the visualization of multiple tables with direct manipulation capabilities and currently work on an {\it overview visualization} that spans the result data space.

\begin{figure}[h]
\includegraphics[width=3.4in]{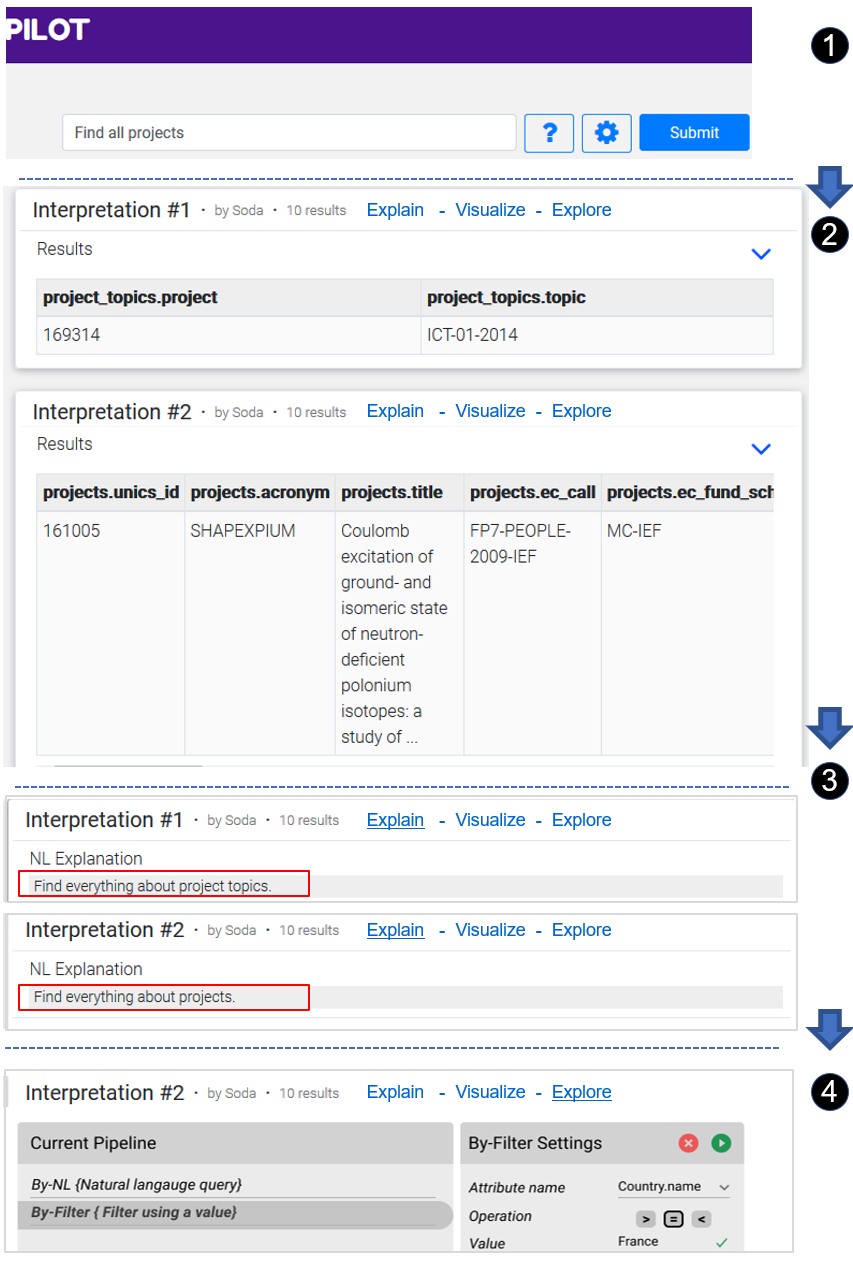}
\caption{Data exploration process in INODE. (1) User starts with an NL question such as \textit{Find all projects.} (2) INODE translates NL question into SQL and returns two different results (interpretations). (3) INODE explains the SQL results in NL. (4) The results can further be explored by pipeline operators such as by-filter to only show projects from France.}
\label{fig:INODE_UI}
\end{figure}

A simple data exploration process with INODE for exploring EU-projects stored in the CORDIS\footnote{https://cordis.europa.eu/} database is shown in Figure \ref{fig:INODE_UI}. 

\section{Open Research Challenges Beyond INODE}
\label{sec:tomorrow}
A full-fledged data exploration system should {\it learn about data sources}, {\it learn about users and queries},  and leverage this knowledge to facilitate and {\it guide users}.   All these challenges constitute new opportunities for ML research to contribute to DM.

\subsection{Learning about Data Sources}
Traditionally SQL or SPARQL are used to query structured data stored in relational databases and graph databases, respectively. A manual time-consuming data integration process is usually required to integrate new data sources. Additionally, data integration techniques used for structured data are not directly applicable to unstructured data since they do not have a schema. 

There are several data integration challenges that can be tackled with ML. The first one is to {\it automate data integration} \cite{golshan2017data}, in the spirit of existing work that improves entity matching using transformer architectures \cite{brunner2020entity}. The second challenge is to {\it automatically generate knowledge graphs}, in the spirit of neural-network systems such as  Snorkel ~\cite{ratner2017snorkel}. Another powerful concept to automate knowledge graph construction is to combine user dialogs with graph construction \cite{nguyen2017query}. The idea is to augment the knowledge graphs by learning concepts that are commonly queried but do not exist in the graph. In summary, there has been a large amount of research on automatic knowledge base construction. However, the combination of knowledge base construction with natural language query processing has been largely untapped. 

\subsection{Understanding Users and Queries}
Understanding user interests and expertise is a vital component for enabling intelligent data exploration. For instance, the general public interested in black holes has different expectations from an experienced astronomer with a vast knowledge of astrophysics. The challenge is to avoid overwhelming a novice user while providing interesting and relevant information to an expert user. 

The system should constantly improve its behavior by learning and adapting to the user
from task to task. Our operators are a great opportunity to learn and adapt to users, as they provide the ability to choose between utility and novelty, two dimensions that have not been explored together in the past. Additionally, they enable collecting user feedback at the level of individual operators and of a DEP. While ML methods for learning user profiles exist in the context of individual systems for web search or recommendations, they have not been studied before in the context of determining which operator caters for which user in the next step. A simple start is to use regression methods to determine the weight of utility and novelty when exploring data. Additionally, probabilistic language models \cite{DBLP:conf/sigir/CaoGLHC17} or matrix factorization-based approaches \cite{DBLP:conf/www/ZhaoCHC15} can be used to infer users' topical interests, and latent factor models to mine user groups. 

Traditional systems such as SODA, ATHENA,  NALIR, or Logos~\cite{blunschi2012soda, saha2016athena, Li:2014:NALIR, logos} use pattern-based approach for NL to SQL or SQL to NL, or supervised ML~\cite{kim2020NL2SQL, brunner2021valuenet}. 
A new opportunity here is to train a neural network for sequence-to-sequence prediction~\cite{sutskever2014sequence, brunner2021valuenet} for translating from NL-to-SQL and vice versa. 
The key research challenge is how to use the feedback provided by users to disambiguate queries and feed the gained knowledge back into ML models to improve learning with semantic information for building ML models to tackle disambiguation and context modeling \cite{shekarpour2016question}. 

This should allow to model and solve the two symmetrical translation problems at once. However, ML methods for query translation typically require large amounts of training data. Our vision is that a full-fledged exploration system should be able to leverage both pattern-based and ML-based approaches to provide the most relevant answers to the user. 

\subsection{Generating Data Exploration Pipelines}
Understanding queries and users serves the ability to provide guidance in generating DEPs.
This challenge can be approached in different ways depending on the user’s expertise and willingness to provide feedback. In a scenario where a DEP is given (see example in Section~\ref{sec:astro}), the problem could be cast as finding the right parameters for each query in the DEP. In a scenario where the user is providing the next query, it could be seen as a query completion problem. In a scenario where the user does not write exploration queries and only provides feedback on results, it could be seen as the problem of learning the user’s DEP. All these cases result in partially-guided or fully-guided exploration.

Furthermore, since DEPs bring together several data access modalities, which may be initiated by the user (e.g. a user query) or by the system (recommendations or explanations), the system needs to learn how to use its options to help the user in meaningful and unobtrusive
ways. While there has been work on each of these capabilities individually (e.g., recommendations
or query explanations), these efforts only focus on small parts of the problem lacking a holistic understanding of the behavior and dynamics of a multi-aspect system.
ML approaches and in particular, Reinforcement Learning (RL) and Active Learning (AL) can be leveraged.

\emph{Partial Guidance with AL}. AL is claimed to be superior to faceted search when the goal is to help users formulate queries. Systems like AIDE~\cite{DBLP:journals/tkde/DimitriadouPD16} and REQUEST~\cite{DBLP:conf/bigdataconf/GeXLSC16} assist users in constructing accurate exploratory queries, while at the same time minimizing the number of sample records presented to them for labeling. Both systems rely on training a decision tree classifier to build a model that classifies unlabeled records. A bigger challenge is to leverage AL in generating and refining queries that go beyond SQL predicates. 

\emph{Full-Guidance with RL}. RL and Deep RL are becoming~\cite{DBLP:conf/sigmod/MiloS20} the methods of choice for creating exploration pipelines and for generating DEPs based on a simulated agent experience~\cite{DBLP:conf/sigmod/ElMS20,DBLP:journals/pvldb/SeleznovaOAS20}. In~\cite{DBLP:conf/sigmod/ElMS20}, a Deep RL architecture is used for generating notebooks that show diverse aspects of a data set in a coherent narrative. In~\cite{DBLP:journals/pvldb/SeleznovaOAS20}, an end-to-end exploration policy is generated to find a set of users in a collection of groups. Both frameworks accept a wide class of exploration actions and do not need to gather exploration logs. An open question is the applicability of this framework to specific data sets and the transferability of learned policies across data sets.

\subsection{Evaluating Data Exploration Pipelines}
Evaluating data exploration requires a holistic approach that addresses {\it performance}, {\it quality} and {\it user experience} aspects~\cite{DBLP:conf/sigmod/EichmannZBK20}. The iterative multi-step nature of DEPs makes evaluation particularly challenging since it is not simply a matter of summing up multiple local evaluations at each step. Regarding performance, the challenge lies in the design of logging mechanisms to export the performance of different steps in DEPs. Such logging must capture in fine details the time and memory usage each step takes so that they can be aggregated to assess a full DEP. The aggregation must be carefully designed to include user actions such as exploring multiple paths and backtracking. While some users prefer the shortest path to a goal, others may be more interested in exploring different paths. 

Multiple evaluation questions arise: “How good are DEPs? How good are NL to SQL/SPARQL translations and vice versa? How good are the next-operator or next-data recommendations and explanations?” The novelty of our approach lies in the ability to {\it jointly assess performance, quality and user experience} for each use case.

We categorize INODE's evaluation metrics into {\it system metrics} and {\it human factors}~\cite{DBLP:journals/vldb/RahmanJN20}. 
To answer above questions, certain key challenges need to be answered. The first challenge lies in extracting metrics from user interactions. We are addressing this by designing logging mechanisms to export various parameters. We use latency and memory usage for evaluating the system performance of a DEP. We also record the number of clicks, interaction time and user feedback. While some users prefer the shortest path to a goal, others may be more interested in exploring different paths.

Another challenge lies in the iterative multi-step nature of DEPs since it is not simply a matter of summing up multiple local evaluations at each step. An additional key challenge is human subject bias. To avoid such biases, our initial studies are designed in multiple stages such as pre-qualification of users, randomly assigning users to different treatment groups and finally feedback from users.
  
User acceptance of our system is strongly related to two key factors: {\it accuracy} - which reflects the closeness of exploration results to desired results, and {\it controllability} - which reflects the ability of our system to guide the user in the exploration~\cite{DBLP:conf/sigmod/EichmannZBK20}. Accuracy is computed using standard methods such as precision and recall, and controllability is the inverse of the number of user-interactions. Our first endeavor is to deploy user studies that explore the relationship between accuracy and controllability. To do so, we will perform factorial design analysis and deploy questionnaires for exploration scenarios we are defining with our use case providers.

The key idea of using ML techniques here is to capture the dependence on users and data and learn different exploration profiles. A major challenge is to design pilot studies in a sound manner with direct observations via questionnaires and indirect ones (such as mouse tracking) to generate labeled datasets for learning. Using ML techniques, and in particular ensemble learning and multi-task learning, constitutes an unprecedented opportunity to adapt the evaluation to users and data.

\section{Acknowledgements}
This project has received funding from the European Union’s Horizon 2020 research and innovation program under grant agreement No 863410. 
\bibliographystyle{ACM-Reference-Format}
\bibliography{rec-bibliography}

\end{document}